\pdfoutput=1

\documentclass[11pt]{article}

\usepackage{booktabs}
\usepackage{tablefootnote}
\usepackage{multirow}
\usepackage[table]{xcolor}
\usepackage{svg}
\usepackage{tcolorbox}

\usepackage{siunitx}
\tcbuselibrary{listings, skins, breakable, theorems}

\newtcolorbox[auto counter]{responsebox}[2][]{%
  float=htb,
  title={Box~\thetcbcounter: #2},
  label={},
  colback=bluebg,
  colframe=blue!60!black,
  coltitle=white,
  fonttitle=\bfseries,
  fontupper=\small\ttfamily,
  boxrule=0.5mm,
  rounded corners,
  width=\textwidth,
  enhanced,
  breakable,
  #1 
}

\newtcolorbox[auto counter]{responsebox*}[2][]{%
  float*=htb,
  title={Box~\thetcbcounter: #2},
  label={},
  colback=bluebg,
  colframe=blue!60!black,
  coltitle=white,
  fonttitle=\bfseries,
  fontupper=\small\ttfamily,
  boxrule=0.5mm,
  rounded corners,
  width=\textwidth,
  enhanced,
  breakable,
  #1
}

\expandafter\def\csname tcb@cnt@responsebox*autorefname\endcsname{Box}

\usepackage[preprint]{acl}

\usepackage{times}
\usepackage{latexsym}

\newcommand{\appref}[1]{\hyperref[#1]{Appendix~\ref*{#1}}}

\usepackage[T1]{fontenc}

\usepackage[utf8]{inputenc}
\usepackage{adjustbox}

\usepackage{microtype}

\usepackage{inconsolata}

\usepackage{graphicx}
\definecolor{answer_attempt}{HTML}{2c7bb6}
\definecolor{hedge}{HTML}{abd9e9}
\definecolor{clarification}{HTML}{fdae61}
\definecolor{refuse}{HTML}{d7191c}

\definecolor{gptfour}{HTML}{018571}
\definecolor{qwen}{HTML}{80cdc1}     
\definecolor{deepseek}{HTML}{a6611a}     
\definecolor{gptmini}{HTML}{dfc27d}
\definecolor{llama}{HTML}{9467bd}
\definecolor{dpollama}{HTML}{8c564b}

\definecolor{bluebg}{HTML}{E2ECF6}

\newcommand{\granu}{\raisebox{-0.1em}{\includegraphics[height=1.3em]{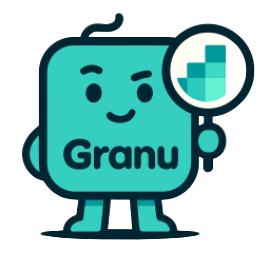}}}

\title{\granu\hspace{0.1em}Granuscore: A Reference-Free Measure of Granularity for Text Analysis and Question Answering}

\author{Lukas Ellinger, Alexander Fichtl, Miriam Anschütz, and Georg Groh \\
  School for Computation, Information and Technology \\
  Technical University of Munich, Germany \\
  \texttt{\{\href{mailto:lukas.ellinger@tum.de}{lukas.ellinger}, miriam.anschuetz, alexander.fichtl\}@tum.de, grohg@cit.tum.de}}
\date{}

\usepackage{enumitem}
\usepackage{makecell}

\begin{document}
\maketitle
\begin{abstract}
Natural language conveys information at varying levels of granularity, from fine-grained references to broad descriptions. While granularity is fundamental to human communication, existing measures mostly capture surface detail or sentence specificity. 
We introduce \textbf{Granuscore}, a reference-free measure of granularity that leverages structural properties of a hierarchical embedding space. Granuscore reliably recovers hierarchical orderings on the \textsc{Granola-EQ} dataset and captures expected differences in granularity across discourse contexts. Across domains, we further show that Granuscore explains non-linear variation in sentence specificity beyond sentence length. Finally, we apply Granuscore to four question-answering benchmarks and analyze how granularity differs for questions, gold answers, and model outputs across response outcomes. The analysis reveals consistent differences in model behavior and provides a principled lens for characterizing the difficulty of QA datasets. Together, the results position Granuscore as a scalable, broadly applicable tool for analyzing granularity in text.
\end{abstract}

\section{Introduction}
\begin{figure}[t]
    \includegraphics[width=\linewidth]{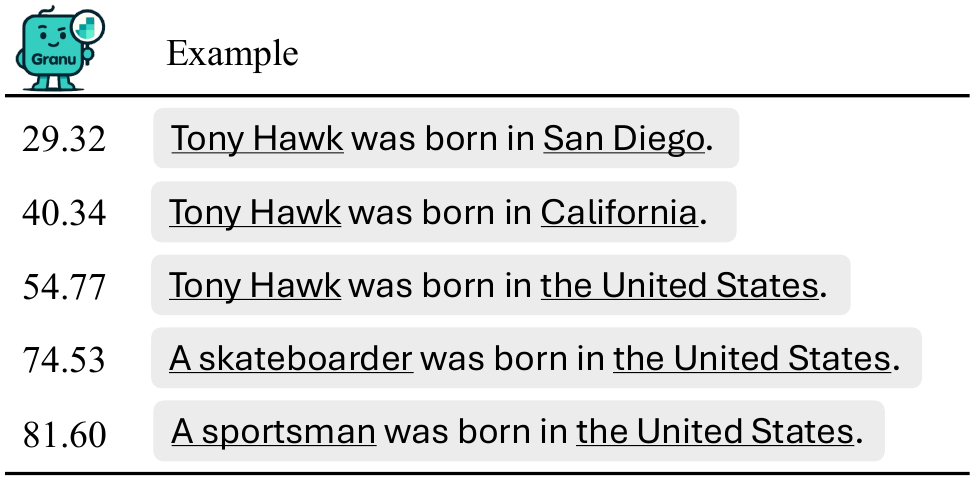}
    \setlength{\belowcaptionskip}{-10pt}
    \vspace*{-5mm}
    \caption{Sentences with referential units varying in granularity. Units that differ across sentences are underlined. Replacing fine-grained terms with coarser alternatives increases sentence granularity: lower Granuscores indicate finer expressions.}
    \label{fig:abstract}
\end{figure}

Natural language varies not only in \emph{what} information is conveyed, but also in \emph{how coarsely or finely} that information is expressed. Consider the sentences in \autoref{fig:abstract}. A speaker may refer to a person as \emph{Tony Hawk}, \emph{a skateboarder}, or \emph{a sportsman}, and may locate an event in \emph{San Diego}, \emph{California}, or \emph{the United States}. These alternatives preserve the underlying fact while referring to it at different levels. We refer to this dimension as \emph{granularity}: the level of abstraction at which entities or events are represented in language \citep{mulkar-mehta_granularity_2011, rosch_basic_1976, hobbs_granularity_1985}.

Granularity is not incidental: speakers adapt the level of abstraction of their descriptions depending on conversational context and task requirements \citep{mulkar-mehta_granularity_2011, hobbs_granularity_1985}. When uncertain, speakers often prefer coarser fact descriptions that remain informative without overcommitting. Conversely, when common ground is established, more fine-grained references become appropriate \citep{yona_narrowing_2024}. Granularity should therefore be understood as a deliberate strategy that balances reliability and audience expectations.

Prior work suggests that linguistic granularity affects how information is perceived and used. In dialogue systems, too fine-grained or coarse responses can reduce user satisfaction \citep{adiwardana_towards_2020, thoppilan_lamda_2022}. Similarly, in simplified language settings, controlling granularity is important for accessibility and comprehension as it reduces the cognitive load \citep{oecd_adults_2024, anschutz_german4all_2025}. However, studying such effects systematically is difficult because existing approaches do not provide a scalable, reference-free measure of granularity at the sentence level.

Our contributions are as follows:
\begin{itemize}[noitemsep, topsep=2pt]
    \item We introduce \textbf{Granuscore}, a reference-free measure of granularity that exploits structural properties of a hierarchical embedding space.
    \item We validate Granuscore both empirically and conceptually. It reliably recovers human-annotated orderings on \textsc{Granola-EQ}~\citep{yona_narrowing_2024} and captures expected granularity differences across discourse contexts.
    \item We show that across domains Granuscore explains non-linear variation in sentence specificity beyond sentence length.
    \item We demonstrate the practical relevance of Granuscore for question answering. Evaluating six language models on four QA benchmarks, we identify consistent differences in granularity between questions, gold answers, and model outputs across response outcomes. These patterns provide a principled lens for characterizing QA dataset difficulty and analyzing model behavior.
    \item We release Granuscore as a \href{https://github.com/lukasellinger/granuscore}{pip package} to ensure reproducibility and enable its usage for further research or production.
\end{itemize}


\section{Background and Related Work}
\paragraph{Granularity}
\citet{mulkar-mehta_granularity_2011} describe granularity in natural language as shifts between coarse and fine descriptions, where higher-level representations abstract from more detailed components. Related perspectives appear in cognitive science, where concepts are organized at different levels of abstraction within taxonomies \citep{rosch_basic_1976}. Further, foundational work by \citet{hobbs_granularity_1985} argues that intelligent reasoning requires representing the world at multiple levels of granularity and switching between them as needed, allowing complex phenomena to be modeled through simpler abstractions.

A related property is \emph{term specificity}, which refers to identifying index terms distinguishing one class of documents from others. In particular, \citet{kim_relationship_2006} describe \emph{hierarchical specificity} as a term’s position within a generic–specific hierarchy, where narrower terms correspond to more specific concepts, matching the notion of granularity.

We capture these ideas using structural properties of a hierarchical embedding space. Unlike approaches relying on manually constructed hierarchies, this enables estimating granularity without being restricted to predefined vocabularies.

\paragraph{Sentence Specificity}
\emph{Sentence specificity} refers to the extent to which a sentence conveys concrete information and supports consistent interpretation across readers \citep{li_improving_2016, ko_domain_2019}. Prior work has shown its relevance for reading comprehension \citep{dixon_processing_1987} and establishing common ground in dialogue \citep{djalali_modeling_2011}.

Although finer-grained references often increase sentence specificity, granularity and sentence specificity capture different properties. Sentence specificity reflects the amount of descriptive information conveyed by a sentence, whereas granularity describes the level at which referential expressions occur within a semantic hierarchy. A sentence can therefore become more specific by adding descriptive details without changing the granularity of its referents. For example, \textit{``The skateboarder won the competition''} becomes more specific in \textit{``The skateboarder won the competition and set a new record.''}. The referents remain at the same granularity level, but the sentence conveys more information.

\paragraph{Granularity Evaluation}
While granularity has been implicitly discussed in work on specificity, informativeness, and semantic hierarchies \citep{thoppilan_lamda_2022, adiwardana_towards_2020, ko_domain_2019, li_improving_2016}, existing automatic evaluations typically rely on taxonomy depth (e.g., WordNet hypernym levels \citep{miller_wordnet_1994} or hierarchical relations in knowledge graphs such as Wikidata \citep{vrandecic_wikidata_2014, huang_can_2023}). However, these approaches require entities to exist in the underlying taxonomy and therefore provide limited coverage for free-form text. In contrast, embedding-based approaches can operate directly on arbitrary text.

\citet{huang_can_2023} propose an automatic benchmark for measuring specificity using transitive relations derived from Wikidata. However, the induced orderings can yield unintuitive comparisons, for example, ranking \emph{Mexico} as more granular than \emph{Colombia}, or \emph{historian} as more granular than \emph{writer}. We therefore acknowledge this dataset but refrain from using it in our experiments.

\citet{yona_narrowing_2024} introduce \textsc{Granola-EQ}, a question answering dataset with explicitly controlled answer granularity levels. They show that standard decoding methods tend to produce overly granular and often incorrect answers. We build on this dataset to train Granuscore and extend their analysis by applying granularity estimation to a broader set of QA datasets, studying how granularity relates to model outputs, correctness, and dataset difficulty.

\paragraph{Training Signals for Informativeness and Interestingness}
The informativeness of model responses plays a central role in user engagement and response quality \citep{adiwardana_towards_2020, thoppilan_lamda_2022}. While early work relies on human annotation to supervise informativeness \citep{adiwardana_towards_2020, thoppilan_lamda_2022}, more recent approaches use LLM-based judges to obtain relative preference signals by comparing response pairs \citep{wu_balancing_2025}. Relatedly, \citet{onozeki_enhancing_2025} introduce interestingness as a training signal and assign scores using an LLM judge. 

In contrast to these approaches, which depend on human supervision or pairwise or model-based judgments, Granuscore provides a reference-free, scalable signal that measures granularity on an absolute and interpretable scale.

\section{Granuscore}
\begin{figure}[t]
    \includegraphics[width=\linewidth]{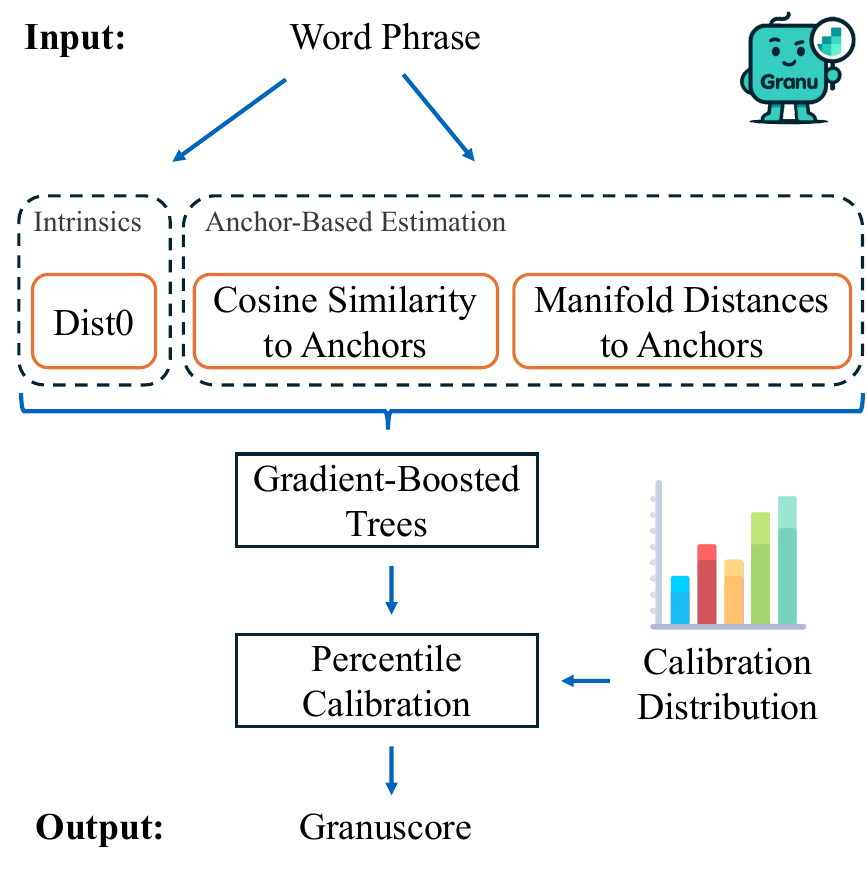}
    \setlength{\belowcaptionskip}{-10pt}
    \vspace*{-5mm}
    \caption{\textbf{Granuscore pipeline}: extraction of hierarchical depth (\textit{Dist0}) and comparison to anchor entities, followed by gradient-boosted trees and percentile calibration to produce a scalar granularity score.}
    \label{fig:pipeline}
\end{figure}

Granuscore measures semantic granularity by exploiting structural properties of a hierarchical embedding space, where \textbf{lower scores} correspond to \textbf{finer-grained expressions}. We build on the \href{https://huggingface.co/Hierarchy-Transformers/HiT-MiniLM-L12-WordNetNoun}{\textsc{Hierarchy Transformer model}} (HiT) proposed by \citet{chen_language_2024}, who train transformer encoders to represent hierarchical structure in a hyperbolic embedding space modeled as a Poincaré ball. In this geometry, hierarchical relations are represented by radial distance from the origin: more specific concepts lie farther from the center, while more general concepts lie closer. We denote this radial distance as \textit{Dist0}, which captures hierarchical depth and serves as a primary signal for granularity. We use the variant trained on the WordNet hierarchy, as WordNet \citep{miller_wordnet_1994} provides a broad-coverage commonsense structure.

While \textit{Dist0} captures global hierarchical position, additional signals can be obtained by relating to other entities in the space. Therefore, we compare the input embedding against a set of anchor entities and derive features from the resulting pairwise relations. In our default configuration, we use 999 randomly sampled fixed anchors, which performed best in our ablation (\autoref{app:k-ablation}). Alternative strategies are described in \autoref{sec:methods}.

\autoref{fig:pipeline} illustrates the resulting pipeline. Given an input word or phrase, the model first obtains a hierarchical embedding and extracts \textit{Dist0}. It then computes pairwise similarity and distance features to the anchor entities using a \href{https://huggingface.co/datasets/philippesaade/wikidata}{Wikidata-derived} embedding index. To map these features to a scalar granularity score, we train gradient-boosted decision trees using \textsc{LightGBM} \citep{ke_lightgbm_2017}. The model operates directly on the raw similarity and distance values, allowing it to capture fine-grained interaction patterns that would be lost under pre-aggregation. Details on the training procedure and model hyperparameters are provided in \autoref{app:model-params}. Because the resulting raw scores depend on the annotations of \textsc{Granola-EQ}, we convert them to percentile scores using a fixed calibration distribution. We choose the WordNet noun set (approximately 119k concepts), which was also used to train the HiT model, providing an annotator-independent alignment. \appref{app:granuscore-labeled-levels} shows how annotation levels map to raw and percentile scores.

\subsection{Dataset}
\label{sec:dataset}
To train the \textsc{LightGBM} model, we use \textsc{GRANOLA-EQ} \citep{yona_narrowing_2024}, an extension of the \textsc{ENTITYQUESTIONS} dataset \citep{sciavolino_simple_2021}. Each dataset entry consists of a question and a set of answers referring to the same underlying \emph{reference entity} at different levels of granularity. We refer to the ordered list of such answers as an \emph{answer hierarchy}, and to the individual answers as \emph{granularity realizations}.

During preprocessing, we remove entries with more than four granularity realizations (fewer than 1.2\% of the data) as these typically reflect inconsistencies introduced during generation. The resulting dataset contains, on average, approximately three realizations per question (2\% with one, 22\% with two, 62\% with three, and 14\% with four).

Since \textsc{GRANOLA-EQ} was generated by prompting an LLM to list increasingly coarse answers, the number of realizations per question varies, and no fixed hierarchical structure is enforced (e.g., \textit{city} $\rightarrow$ \textit{state} $\rightarrow$ \textit{country}). The LLM implicitly determines the resolution of the answer hierarchy it considers appropriate for a given question. To obtain comparable training targets, we normalize the answer levels to a continuous scale from 1 (most fine-grained) to 4 (most coarse-grained); for example, a hierarchy with three answers is mapped to levels ${1, 2.5, 4}$.

Due to the construction of \textsc{GRANOLA-EQ}, the same entity may appear at different granularity levels across dataset entries depending on the question context (e.g., \textit{England} appears 487 times with a mean granularity of 3.25 and variance 0.44).
We retain these variations, allowing the model to learn from multiple granularity annotations of the same realization and encouraging generalization.

Finally, to prevent data leakage, we enforce that no granularity realization appears in more than one split. The final dataset consists of 6,702 training samples and 1,220 samples each for development and test. For dataset details, see \autoref{app:granola-dataset}.

\subsection{Extension to Multi-Word Inputs}
Because Granuscore is defined for individual referential units, we extend it to sentences and longer text spans by decomposing the text. We use a \textsc{spaCy}-based splitter \citep{honnibal_spacy_2020}. 
Noun phrases are kept intact, while stop words and non-informative symbols are removed. This preserves referential expressions that convey granularity while avoiding fragmentation of multi-word concepts. If no referential units can be identified (e.g., the input consists solely of stop words), we assign a Granuscore of 100, corresponding to the coarsest granularity score.

We avoid decomposing inputs into atomic facts, as a single fact may contain multiple entities with different granularity levels, making fine-grained attribution difficult. For example, in \autoref{fig:abstract}, both the person and the location contribute to the sentence’s granularity, such that modifying either referent changes the perceived granularity. Moreover, atomic fact decomposition can introduce or duplicate lexical material not explicitly present in the original sentence, which may bias the resulting scores \citep{wanner_dndscore_2024}.

Finally, we compute the granularity score for a multi-word input using a two-step aggregation. First, we compute a sentence-level Granuscore by averaging the scores of the extracted referential units within each sentence. We then aggregate across sentences by taking the mean of the bottom 80\% of sentence Granuscores, which reduces the influence of unusually high scores. We evaluate a range of alternative aggregation strategies and compare them in \autoref{app:aggregation_strategy}. Based on this ablation, we adopt this aggregation as the default, as it shows the strongest performance.

\subsection{Methods}
\label{sec:methods}

\begin{figure}[t]
    \includegraphics[width=\linewidth]{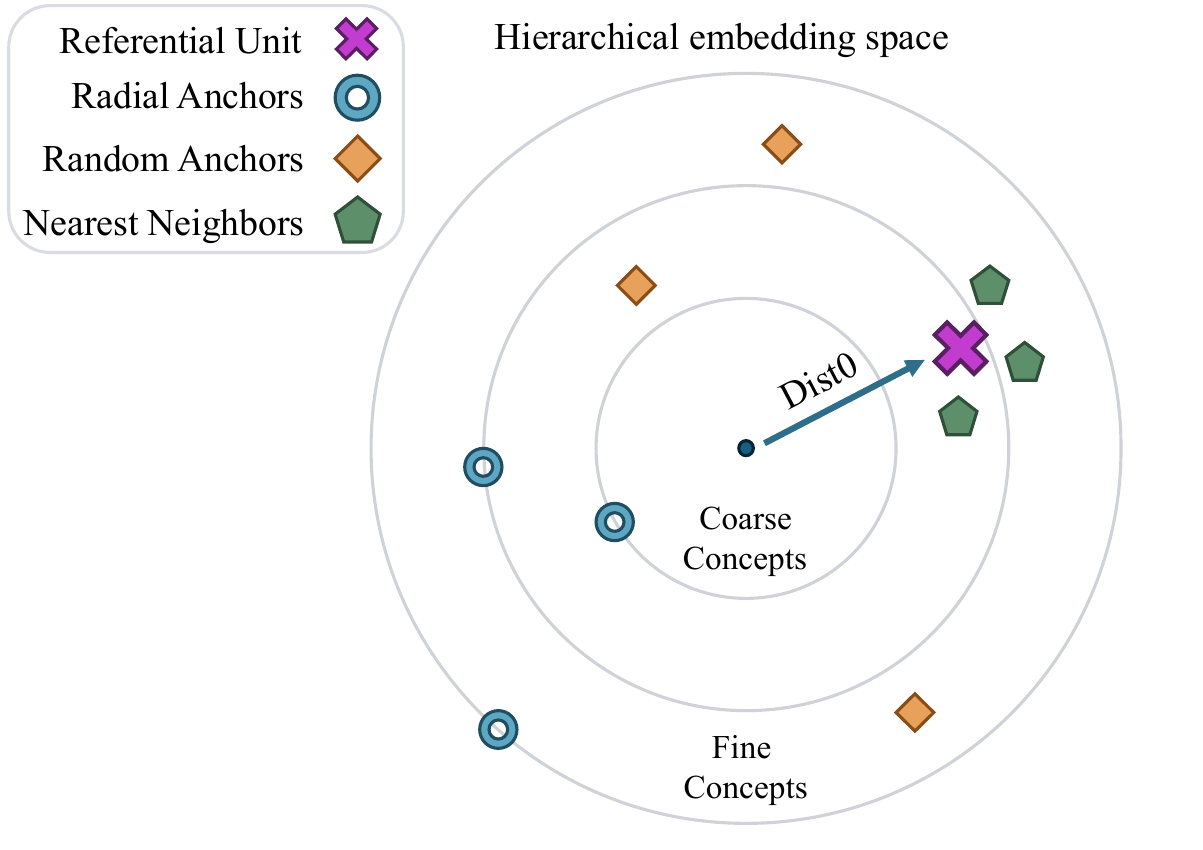}
    \setlength{\belowcaptionskip}{-10pt}
    \vspace*{-5mm}
    \caption{Illustration of the hierarchical embedding space. Referential units are embedded in a radial semantic hierarchy, with coarser concepts closer to the center and finer-grained concepts in outer regions.}
    \label{fig:embedding-space}
\end{figure}

To evaluate the effectiveness of Granuscore, we compare it against several baselines and variants that estimate granularity using lexical, hierarchical, or embedding-based signals. The embedding-based variants are illustrated in \autoref{fig:embedding-space}.

\begin{itemize}[leftmargin=*,noitemsep]
\item \textbf{Word Count:} Number of words in the text. Negative word count so that higher scores correspond to coarser concepts.
\item \textbf{WordNet Hierarchy:} Average depth of mapped WordNet synsets; deeper nodes correspond to finer concepts.
\item \textbf{GPT-4.1 mini:} Few-shot prompting to estimate granularity (\autoref{app:llm-prompt}).
\item \textbf{HiT Dist0:} radial distance \textit{dist0} only.
\item \textbf{Nearest Neighbors (NN):} top-$k$ cosine-similar entities as anchors.
\item \textbf{Random:} $k$ dynamically sampled anchors.
\item \textbf{Random Anchors:} fixed $k$ random anchors.
\item \textbf{Radial Anchors:} fixed set of $k$ anchors sampled across HiT \textit{dist0} distance bins.
\end{itemize}
For Nearest Neighbors, Random, and Random Anchors, we evaluate both HiT and MiniLM embeddings. MiniLM serves as a widely used non-hierarchical embedding baseline to contextualize the contribution of hierarchical representations. Radial Anchors are only defined for HiT, as they rely on the HiT \textit{Dist0} radial structure. Additional details on the methods are provided in \autoref{app:neighborhood}. Exact versions of all models used throughout the paper are listed in \autoref{app:models}.

\subsection{Evaluation Approaches}
We evaluate Granuscore across three complementary settings that test different aspects. First, we measure how well methods recover controlled granularity orderings. Second, we examine whether granularity scores capture differences across discourse contexts. Finally, we analyze how Granuscore relates to sentence specificity.

\paragraph{GRANOLA-EQ} 
We first evaluate all methods on the test set of \textsc{GRANOLA-EQ}. We use \textit{Pairwise Accuracy}, defined as the percentage of correctly ordered granularity realizations. 
For realization $R_i$ and $R_j$ with gold ordering $R_{i,\text{gold}} < R_{j,\text{gold}}$, a prediction is considered correct if $R_{i,\text{pred}} < R_{j,\text{pred}}$. 
Pairs with identical gold granularity levels are excluded because the resolution of the \textsc{GRANOLA-EQ} annotations does not define a unique ordering.

We compute this metric in two settings. In the \textit{global} setting, pairwise accuracy is computed across all dataset entries, measuring the ability of a method to assign consistent granularity scores to unrelated entities. This task is particularly challenging because entities may belong to different semantic dimensions and must be placed on a shared granularity scale. For example, a model must compare realizations such as \textit{skateboarder} and \textit{California}, originating from different hierarchies (e.g., \textit{Tony Hawk} $\rightarrow$ \textit{American skateboarder} $\rightarrow$ \textit{skateboarder} $\rightarrow$ \textit{sportsman} and \textit{San Diego} $\rightarrow$ \textit{California} $\rightarrow$ \textit{United States} $\rightarrow$ \textit{America}).

In the \textit{intra-entry} setting, pairwise accuracy is computed within the hierarchy of each entry. This evaluates how well methods recover the local ordering of semantically related realizations.

\paragraph{Discourse Contexts}
Large-scale annotations of granularity for longer text are difficult to obtain, so we instead rely on naturally occurring discourse differences as an unsupervised proxy. Scientific papers provide a suitable testbed, as their standardized section structure reflects distinct discourse functions: Introduction sections typically describe the broader research context using more coarse-grained references, whereas Related Work sections contain more fine-grained references to specific prior methods, datasets, and papers, reflecting common rhetorical structures in scientific writing \citep{swales1990genre,day2012write}.

We apply Granuscore to scientific articles from the \textsc{S2ORC} corpus \citep{lo_s2orc_2020}. We sample 1{,}000 papers and compare the granularity of Introduction and Related Work sections. Further details on the sampling and filtering procedure are provided in \autoref{app:papers}.

\paragraph{Sentence Specificity}
Finally, we examine how Granuscore relates to sentence specificity. We analyze human-annotated datasets from \citet{ko_domain_2019} and \citet{li_improving_2016}. The former covers movie reviews, tweets, and Yelp reviews, while the latter contains sentences from news articles.

Sentence length (word count) is a strong baseline predictor with Spearman correlations of 0.45 (Twitter), 0.58 (movie reviews), 0.68 (Yelp), and 0.67 (news). We therefore quantify Granuscore’s contribution to sentence specificity beyond sentence length by fitting Generalized Additive Models (GAMs). This allows us to isolate the contribution of each predictor via explained deviance.

\section{Results}
Below, we report results for the three evaluation settings introduced above.
\subsection{GRANOLA-EQ}
\label{res-granola-eq}
\autoref{tab:lgb_results} reports the performance of all methods on \textsc{GRANOLA-EQ} using Global and Intra-entry Pairwise Accuracy and Exact Ordering Accuracy. Additional metrics are provided in \appref{app:additional_metrics_granola}.

Across all methods, intra-entry pairwise accuracy consistently exceeds global pairwise accuracy, indicating that ranking realizations within the same semantic hierarchy is easier than assigning consistent scores across unrelated entities. Exact ordering accuracy is consistently lower, reflecting the greater difficulty of recovering the full hierarchy rather than individual pairwise relations.

The radial depth signal already provides a strong baseline: HiT Dist0 achieves 80.82\% global pairwise accuracy and 87.86\% intra-entry accuracy. 

\begin{table}[t]
\small
\setlength{\tabcolsep}{4pt}
\centering
\begin{tabular}{l@{\hspace{2pt}}ccc}
\toprule
Method & \makecell{Global PW\\Acc.} & \makecell{Intra PW\\Acc.} & Exact \\
\midrule
Word Count & 50.49 & 51.54 & 28.80 \\
WordNet$^\dagger$ & 58.12 & 67.32 & 60.39 \\
GPT-4.1 mini & 76.76 & 81.58 & 58.21 \\
HiT Dist0 & 80.82 & 87.86 & 73.50 \\
\midrule
MiniLM NN & 67.40 & 69.50 & 48.97 \\
MiniLM Random & 67.55 & 70.57 & 48.80 \\
MiniLM RandomAnch & 67.45 & 71.15 & 49.15 \\
\midrule
HiT NN & 81.79 & 86.47 & 70.17 \\
HiT Random & 80.80 & 84.74 & 69.32 \\
HiT RadialAnch & 82.31 & 88.35 & 71.71 \\
HiT RandomAnch & 83.00 & 88.15 & 72.48 \\
\midrule
HiT Dist0 + NN & 82.86 & 87.02 & 71.54 \\
HiT Dist0 + Random & 82.01 & 86.82 & 70.85 \\
HiT Dist0 + RadialAnch & \textit{83.22} & \textit{88.83} & \textit{73.85} \\
HiT Dist0 + RandomAnch & \textbf{83.76} & \textbf{89.03} & \textbf{74.36} \\
\bottomrule
\end{tabular}
\setlength{\belowcaptionskip}{-10pt}
\caption{Comparison of methods on the \textsc{GRANOLA-EQ} test set. We report Global Pairwise Accuracy (PW Acc.), Intra-entry Pairwise Accuracy (Intra PW Acc.), and Exact Ordering Accuracy (Exact). Bold indicates the best result and italic indicates the second-best result across methods. $^\dagger$: WordNet could only derive a granularity level for 17.61\% of the realizations.}
\label{tab:lgb_results}
\end{table}

In comparison, anchor-based HiT methods generally outperform it in global pairwise accuracy, with HiT Random Anchors outperforming HiT Dist0 by 2.18 percentage points. Notably, anchors sampled across the embedding space outperform nearest neighbors, suggesting that global structure is more informative for estimating granularity than local similarity.

Combining HiT Dist0 with Random Anchors achieves the highest scores across all metrics. Compared to HiT Dist0, it improves by +2.94 (global) and +1.17 (intra-entry) points, and over HiT Random Anchors by +0.76 and +0.88 (bootstrap resampling, $N=20{,}000$, $p<0.002$ global; $p<0.05$ intra-entry). 

In contrast, all MiniLM-based variants perform substantially worse (best global: 67.55\%) and yield nearly identical scores regardless of the anchor selection strategy. This suggests that the underlying embedding geometry provides a weaker signal for granularity than HiT.

Among additional baselines, the WordNet hierarchy achieves 58.12\% global pairwise accuracy despite covering only 17.61\% of realizations. This shows that lexical taxonomies contain meaningful signals for granularity when applicable, but their use is limited by coverage. GPT-4.1 mini performs competitively in pairwise ordering but shows lower exact ordering accuracy. Here, manual inspection indicates that the model frequently assigns identical granularity levels to multiple realizations, thereby reducing its ability to recover the full hierarchy. Finally, the Word Count baseline performs close to random (50\%), confirming that granularity is not reflected in sentence length.

Overall, these results show that granularity is best captured by combining hierarchical depth and anchor comparisons in the embedding space.

\subsection{Granuscore Across Paper Sections}
\label{res-papers}
Beyond the gold-labeled setting, we evaluate whether Granuscore captures differences across discourse contexts.
Across the sampled papers, 68.71\% of paired comparisons exhibit a higher Granuscore for the Introduction than for the Related Work section, indicating that Introduction sections tend to use more coarse-grained language.
This difference is highly significant (paired $t$-test: $p\leq5.49\times10^{-37}$; Wilcoxon signed-rank test: $p\leq5.73\times10^{-39}$) with a moderate paired effect size ($d_z = 0.42$; \citealp{cohen_statistical_2013}).
Consistent with this ordering, Introduction sections also have a higher average Granuscore ($75.29 \pm 4.94$) than Related Work sections ($72.43 \pm 5.54$).

This pattern aligns with the rhetorical roles of these sections.

\subsection{Correlation to Sentence Specificity}
\label{res-specificity}

\begin{table}[t]
\centering
\setlength{\tabcolsep}{6pt}
\small
\begin{tabular}{lccc}
\toprule
Domain &
\makecell{Expl.\ Dev.\\(Length)} &
\makecell{Expl.\ Dev.\\(Len+Gran)} &
$\Delta$ \\
\midrule
movie    & 0.38 & 0.46 & \textcolor{green!60!black}{+0.08} \\
twitter  & 0.24 & 0.36 & \textcolor{green!60!black}{+0.12} \\
yelp     & 0.52 & 0.56 & \textcolor{green!60!black}{+0.04} \\
news     & 0.45 & 0.55 & \textcolor{green!60!black}{+0.10} \\
\bottomrule
\end{tabular}
\setlength{\belowcaptionskip}{-10pt}
\caption{Explained deviance of generalized additive models (GAMs) predicting sentence specificity. All smooth terms are significant ($p < 2.41 \times  10^{-10}$).}
\label{tab:gam_granularity_specificity}
\end{table}

\begin{figure}[t]
    \centering
    \includegraphics[width=\linewidth]{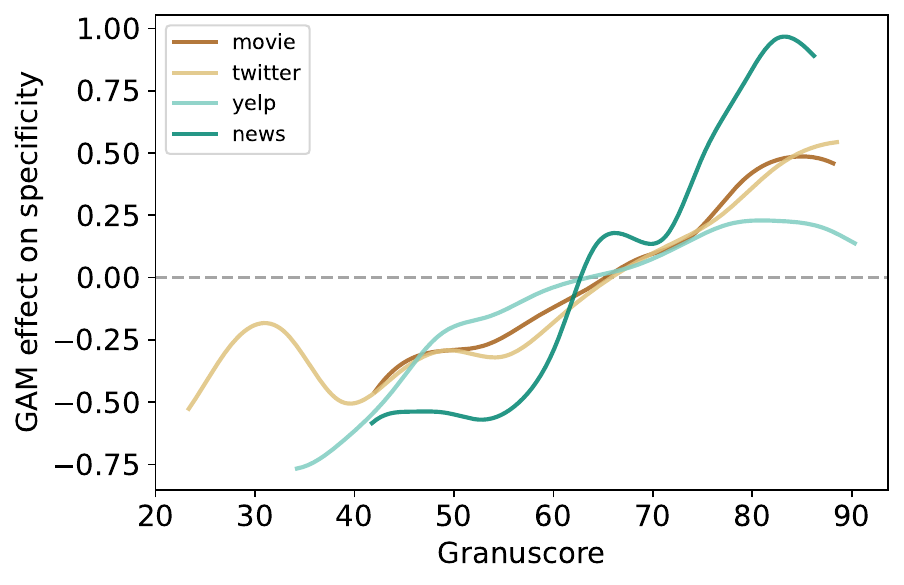}
\setlength{\belowcaptionskip}{-10pt}
\vspace*{-5mm}
\caption{Effect of Granuscore on sentence specificity across domains.
Lower specificity scores correspond to more specific sentences.
The plotted range is restricted to the 1st--99th percentiles of Granuscore to avoid sparse-support regions.}
\label{fig:specificity-gran}
\end{figure}

Finally, we analyze how Granuscore relates to sentence specificity. \autoref{tab:gam_granularity_specificity} shows that adding Granuscore consistently improves explained deviance over a length-only baseline across all domains. Absolute gains range from +0.04 (Yelp) to +0.12 (Twitter), corresponding to relative improvements of 7–50\%. In \autoref{fig:specificity-gran}, we show the estimated effect of Granuscore on sentence specificity. Across all domains, the relationship is non-linear. Lower Granuscore negatively affects the specificity score, indicating an association with more specific sentences. As Granuscore increases, the magnitude of this negative effect decreases. The effect crosses zero between values of roughly 63–66, after which higher scores are associated with less specific sentences. For completeness, the effect of sentence length is shown in \autoref{app:specificity}.

Overall, these results show that Granuscore captures a systematic component of sentence specificity while remaining distinct from it. Although granularity alone does not determine specificity, incorporating Granuscore consistently improves specificity prediction across domains. This pattern aligns with the intuition that references to fine-grained entities tend to appear in more specific sentences, whereas coarse-grained concepts are more common in less specific ones.

\section{Applying Granuscore to QA Datasets}
\label{sec:qa}
We apply Granuscore to several widely used QA datasets to investigate how granularity affects dataset properties and model performance. We use the public splits of \textsc{FACTS Parametric} \citep{cheng_facts_2025} (1{,}047 samples), \textsc{SimpleQA} \citep{wei_measuring_2024} (4{,}255), \textsc{SQuAD} \citep{rajpurkar_squad_2016} (10{,}570), and \textsc{TruthfulQA} \citep{lin_truthfulqa_2022} (817), resulting in a total of 16{,}689 samples.

To relate granularity to model behavior, we evaluate model correctness using \textsc{Qwen3 0.6B}, \textsc{Qwen3-8B}, and \textsc{Qwen3-32B} \citep{yang_qwen3_2025}, \textsc{Olmo 3 7B} \citep{olmo_olmo_2025}, and \textsc{DeepSeek V3.2} \citep{deepseek-ai_deepseek-v3_2025}. These models cover a broad range of model sizes and represent well-established open-weight language models. For \textsc{Qwen3-8B}, we additionally evaluate both standard generation and reasoning-enabled generation (``think’’ mode) to compare performance with and without explicit reasoning.

Model responses are evaluated using GPT-4.1 nano as an LLM-based judge, following the prompt template introduced in \textsc{SimpleQA} \citep{wei_measuring_2024}. Further details are given in \autoref{app:qa-generation}.

\paragraph{Granuscore Gold Answers}
\begin{figure}[t]
    \includegraphics[width=\linewidth]{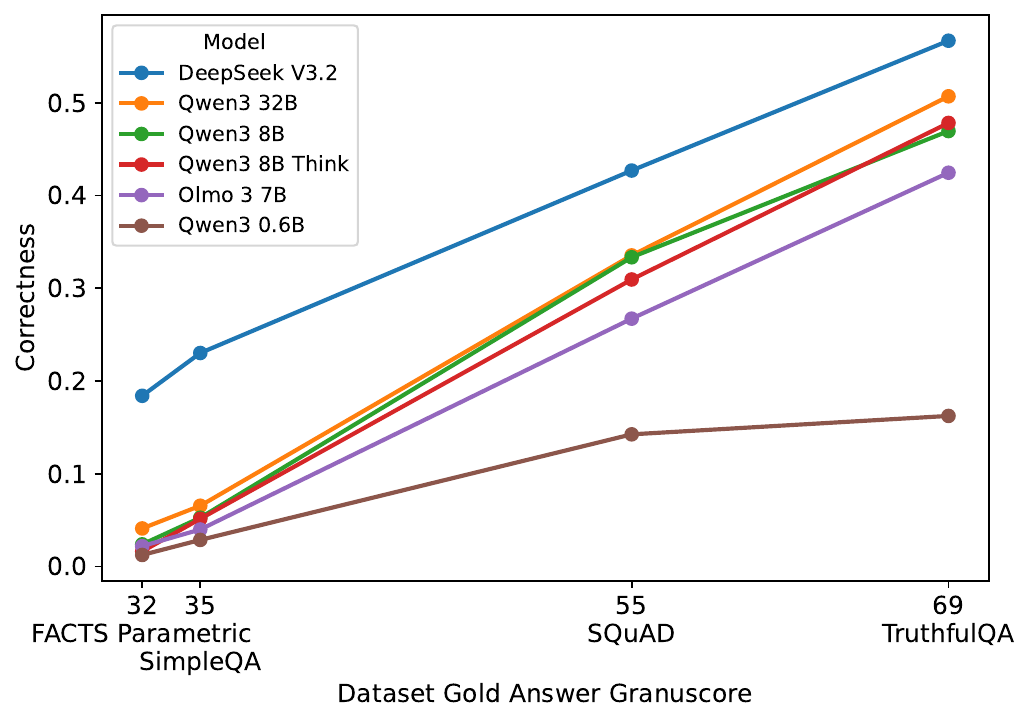}
    \setlength{\belowcaptionskip}{-10pt}
    \vspace*{-5mm}
    \caption{Relationship between dataset-level gold answer Granuscore and model correctness across QA benchmarks. Higher Granuscore datasets are associated with higher correctness across models. All pairwise differences in Granuscore between datasets are statistically significant (Mann–Whitney $U$, $p \leq 1.1 \times 10^{-3}$).}
    \label{fig:qa_datasets_scatter}
\end{figure}

In \autoref{fig:qa_datasets_scatter}, we relate model correctness to the Granuscore of gold answers across QA datasets. For each of the models, correctness varies strongly across datasets, with mean accuracies of 4.99\% on \textsc{FACTS Parametric}, 7.80\% on \textsc{SimpleQA}, 30.24\% on \textsc{SQuAD}, and 43.47\% on \textsc{TruthfulQA}. 

Datasets with lower Granuscores exhibit substantially lower accuracy, while higher Granuscore datasets are associated with improved performance. This pattern is consistent across all evaluated models, suggesting that Granuscore captures a model-independent aspect of question difficulty. Larger models, such as \textsc{DeepSeek V3.2}, achieve consistently higher correctness across datasets, indicating greater knowledge coverage, but follow the same overall trend. In contrast, the smallest model, \textsc{Qwen3 0.6B}, exhibits a weaker slope, likely reflecting general capacity limitations. We observe a similar trend when analyzing the Granuscore of questions (\autoref{app:qa_datasets}). In contrast, potential confounding factors, including answer and question length, word frequency, and syntactic complexity, do not yield comparably consistent relationships with correctness (\autoref{app:qa_datasets}).

\paragraph{Granuscore Across Response Outcomes}
\begin{table}
\centering
\small
\begin{tabular}{lS[table-format=2.1(1.1)]S[table-format=2.1(1.1)]S[table-format=2.1(1.1)]}
\toprule
Type & {Correct} & {Wrong} & {Not Att.} \\
\midrule
Question & 70.1(1.5) & 65.4(0.3) & 67.2(0.6) \\
Gold Answer & 59.4(4.1) & 45.8(0.7) & 48.7(2.7) \\
Answer & 69.6(1.7) & 66.0(1.4) & 72.5(1.3)\\
\bottomrule
\end{tabular}
\setlength{\belowcaptionskip}{-10pt}
\caption{Granuscore (mean $\pm$ std. across models) by response outcome (Correct, Wrong, and Not Attempted). Granuscore distributions differ significantly across outcomes (Mann--Whitney $U$, $p \leq 2.42 \times 10^{-13}$).}
\label{tab:granularity_label_means}
\end{table}

In \autoref{tab:granularity_label_means}, we report mean Granuscore values for questions, gold answers, and model outputs, stratified by response outcome (correct, incorrect, and not-attempted). Model outputs associated with wrong answers exhibit the lowest average Granuscore (66.0), followed by correct answers (69.6), while not-attempted responses show the highest output granularity (72.5). The latter is expected, as abstentions typically consist of general statements indicating an inability to provide an answer.

On the input side, incorrect responses are associated with lower-granularity questions and gold answers, followed by not-attempted cases and then correct responses.

Finally, we analyze the \emph{granularity gap}, defined as the difference between model output and gold-answer Granuscore. The gap is substantially larger for incorrect and not-attempted responses than for correct ones. Using five-fold cross-validation, a logistic regression with granularity gap as the sole predictor achieves an average AUC of 0.62 ($\pm$ 0.005), indicating a moderate, stable association between granularity mismatch and response failure.

\section{Discussion}
\paragraph{Comparison of Methods}
Our results on \textsc{GRANOLA-EQ} highlight the importance of hierarchical structure for estimating granularity. Methods based on HiT consistently outperform approaches using standard sentence embeddings, indicating that granularity is closely tied to hierarchical relations rather than surface similarity. Importantly, the radial depth signal (Dist0) already outperforms several baselines without any training on \textsc{GRANOLA-EQ}, indicating that the hierarchical embedding space itself captures meaningful granularity signals independent of the learned mapping. However, anchor-based comparisons further improve performance, particularly for global ordering. Comparing entities to anchors across the embedding space provides additional relational context, enabling a more reliable and stable estimation of granularity across unrelated hierarchies. Hence, the best results are achieved when combining both signals.

This challenge of comparing unrelated hierarchies is also reflected in the evaluation metrics. We observe a consistent gap between intra-entry and global accuracy: intra-entry comparisons operate within a shared semantic hierarchy (e.g., city $\to$ state $\to$ country), whereas global comparisons require ordering entities from unrelated hierarchies on a common scale. Despite this difficulty, Granuscore provides a strong signal for estimating general granularity across heterogeneous hierarchies.

\paragraph{Correlation to Sentence Specificity}
Further, we show that Granuscore explains non-linear variation in sentence specificity beyond sentence length, which serves as a strong baseline indicator \citep{gao_predicting_2019, ko_domain_2019}. The consistency of this relationship across heterogeneous domains supports the robustness of Granuscore as a general granularity measure.

\paragraph{Granularity and QA Performance}
Our QA analysis case study reveals consistent patterns linking granularity and response outcomes.

First, across all models and 16,689 QA-samples, we observe clear differences in the Granuscore across response outcomes. Questions and gold answers associated with incorrect or not-attempted responses exhibit significantly lower Granuscore values than those associated with correct responses. The effect is particularly pronounced for gold answers, while the difference in question granularity is present but weaker. These findings suggest that Granuscore may serve as a proxy for the difficulty of question–answer pairs and could be incorporated as a signal for deciding when to rely on a model’s internal knowledge versus external tools.

At the dataset level, we observe a complementary trend: datasets with lower granularity, both for gold answers and questions, are substantially harder for models. This suggests that granularity provides a useful lens for characterizing differences in QA-difficulty that are not explained by superficial properties such as answer or question length.

Finally, we observe that incorrect responses tend to exhibit lower output granularity than correct ones. In these cases, models often remain at the level of detail implied by the question rather than adapt their responses to a more appropriate granularity based on their confidence in the answer. This aligns with findings that models struggle to adjust answer granularity \citep{yona_narrowing_2024} and  \citet{kalai_why_2025} arguing that benchmark evaluations incentivize models to guess overly specific answers.

\paragraph{Future Directions}
A natural next step is to use Granuscore as a training signal for language models. Prior work has shown that optimizing for properties such as informativeness and interestingness can improve response quality and user engagement \citep{adiwardana_towards_2020, thoppilan_lamda_2022, onozeki_enhancing_2025}. Similarly, Granuscore could encourage models to generate responses at appropriate levels of granularity. In particular, models could learn to align output granularity with their confidence: when uncertain about fine-grained details, they may respond at a coarser but reliable level (e.g., a broader category or time period). Such behavior mirrors human communication and could help reduce overly fine-grained incorrect answers while preserving informative responses.

Beyond response generation, Granuscore may also support controlled language adaptation, such as simplification, including summarization \citep{stoll_plain_2022} and definition generation \citep{ellinger_simplifications_2025}, where appropriate granularity is crucial for producing accessible yet informative text.

\section{Conclusion}
We introduced Granuscore, a reference-free measure that quantifies the granularity expressed in text using a hierarchical embedding space. Granuscore reliably recovers granularity orderings on the controlled \textsc{GRANOLA-EQ} benchmark, aligns with expected differences across scientific paper sections, and captures non-linear variation in sentence specificity beyond sentence length.

Applied to question answering, Granuscore provides a useful lens for characterizing dataset difficulty and understanding differences in model performance and outputs. 

\section*{Limitations}
\paragraph{Dependence on WordNet Hierarchy.}
Granuscore relies on a single hierarchical embedding model fine-tuned on the WordNet hierarchy. We choose this variant because WordNet provides broad-coverage, general-purpose commonsense structure. This choice might limit granularity estimation in domains that are poorly represented in WordNet. Future work could explore domain-specific hierarchical models and evaluate their impact, whereas we intentionally focus on general applicability in this work.

\paragraph{Human Perception of Granularity.}
While GRANOLA-EQ is manually validated by human annotators, our evaluation does not include a dedicated human study directly comparing Granuscore scores against explicit human granularity judgments. Instead, we focus on broad empirical validation across multiple complementary settings, including hierarchical ordering, discourse-level analyses, sentence specificity, and downstream QA behavior.

\section*{Acknowledgments}
All analysis, research, and ideas are either our own or cited. This work used LLM-based tools for language edits and clarity improvements. This research has been funded by the German Federal Ministry of Research, Technology, and Space (BMFTR) through grant 01IS23069 Software Campus 3.0 (Technical University of Munich) as part of the Software Campus project “Know ELViS”.

\bibliography{main}

\appendix
\section{Example Sentences}
\begin{figure}[t]
    \centering
    \includegraphics[width=\linewidth]{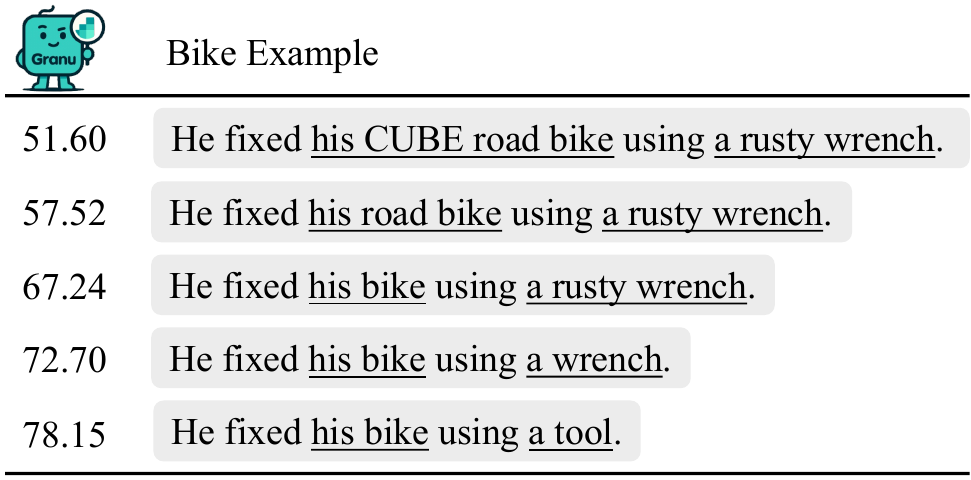}
\caption{Illustration of semantic abstraction. Starting from the specific statement \textit{``He fixed his CUBE road bike using a rusty wrench''}, it can be generalized by abstracting the vehicle and the instrument.}
\label{fig:bike-example}
\end{figure}

\begin{figure}[t]
    \centering
    \includegraphics[width=\linewidth]{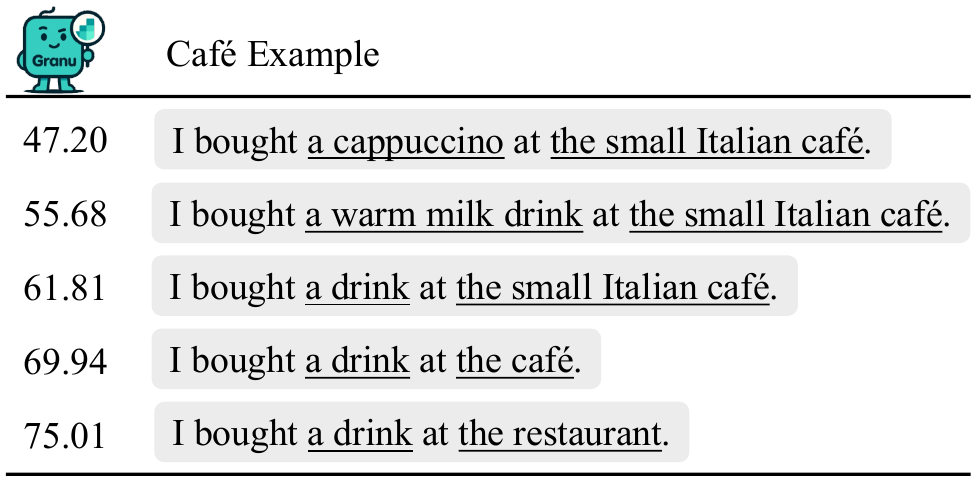}
\caption{Illustration of semantic abstraction. Starting from the specific statement \textit{``I bought a cappuccino at the small Italian café''}, it can be generalized by abstracting the drink type and the venue.}
\label{fig:cafe-example}
\end{figure}

\begin{figure}[t]
    \centering
    \includegraphics[width=\linewidth]{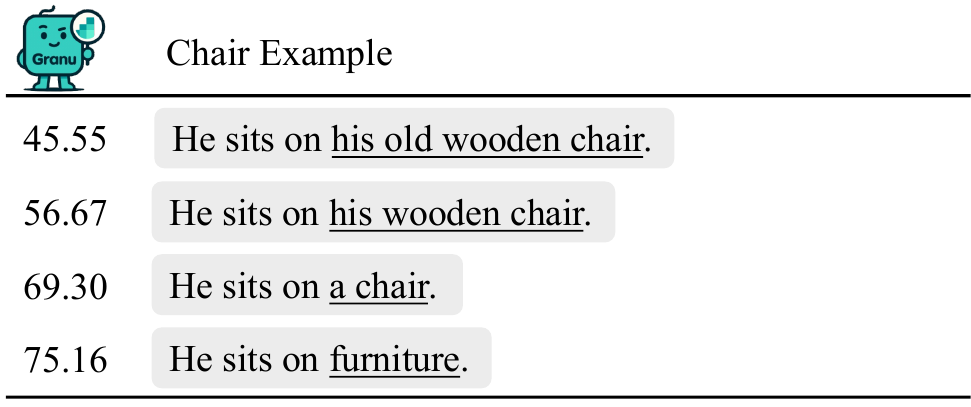}
\caption{Illustration of semantic abstraction. Starting from the specific statement \textit{``He sits on his old wooden chair''}, it can be generalized by abstracting the seating option.}
\label{fig:chair-example}
\end{figure}

We present example sentences with controlled granularity levels together with their assigned Granuscore values in \autoref{fig:bike-example}, \autoref{fig:cafe-example}, and \autoref{fig:chair-example}. They illustrate how the same underlying fact can be expressed at different levels of granularity.

\section{Model Access}
\label{app:models}
To support reproducibility, \autoref{tab:models} lists all models used in this paper, including their names, exact versions, and access providers.

\begin{table*}[t]%
\small
\centering%
\begin{tabular}{@{}l l l@{}}
\toprule%
Name&Version&Access Provider\\%
\midrule%
HiT~\citep{chen_language_2024} & HiT-MiniLM-L12-WordNetNoun & Local\\%
MiniLM~\citep{minilm} & all-MiniLM-L6-v2\tablefootnote{\url{https://huggingface.co/sentence-transformers/all-MiniLM-L6-v2}} & Local\\%
GPT-4.1 nano~\citep{openai2025gpt41} & gpt-4.1-nano-2025-04-14& OpenAI Batch API\\%
GPT-4.1 mini~\citep{openai2025gpt41} & gpt-4.1-mini-2025-04-14& OpenAI Batch API\\%
Qwen3 0.6B~\citep{yang_qwen3_2025} & N/A& Local\\%
Olmo 3~\citep{olmo_olmo_2025} & Olmo-3-7B-Instruct & Local\\%
Qwen3 8B~\citep{yang_qwen3_2025} & N/A& Local\\%
Qwen3 8B Think~\citep{yang_qwen3_2025} & N/A& Local\\%
Qwen3 32B~\citep{yang_qwen3_2025} & N/A& OpenRouter\\%
DeepSeek V3.2~\citep{deepseek-ai_deepseek-v3_2025}& N/A& OpenRouter\\%
\bottomrule%
\end{tabular}%
\caption{Specific model versions used in our experiments. For each model we provide the exact version and the access provider.}%
\label{tab:models}
\end{table*}

\section{Reference Construction}
\label{app:neighborhood}
As the embedding space is unbounded, we approximate its structure using a finite subset of entities. We construct this proxy space from 50,000 randomly sampled Wikidata entities\footnote{https://huggingface.co/datasets/philippesaade/wikidata}. For each entity, we use its title as the textual representation. Wikidata offers broad topical coverage and a relatively clean entity structure, making it a suitable general-purpose semantic reference. We choose an index size of 50,000 entities as a trade-off between computational efficiency and neighborhood fidelity. In preliminary experiments, this size yielded stable neighborhood structures, while larger indices substantially increased runtime (cf. \autoref{app:k-ablation}).

All entity embeddings are indexed using \textsc{FAISS} \citep{douze_faiss_2025}. 
For the Random Anchor and Radial Anchor methods, anchors are sampled once in advance from the index and reused for all queries. For the Nearest Neighbor method, we retrieve the top nearest neighbors using cosine similarity for each query individually. For the Random Neighbor baseline, neighbors are sampled at random from the index for each query.

\section{LLM Prompt}
\label{app:llm-prompt}
We use the following few-shot prompt to annotate granularity levels with an LLM. The prompt includes three example semantic hierarchies comprising a total of 14 realizations together with their expected granularity levels.

\begin{tcolorbox}[
  colback=bluebg,
  colframe=blue!60!black,
  coltitle=white,
  fonttitle=\bfseries,
  fontupper=\small\ttfamily,
  boxrule=0.5mm,
  rounded corners,
  title={User Prompt: LLM Granularity}
]
You are an expert annotator for granularity.\\

Your task is to assign a granularity score to an answer using a 4-point Likert scale.
Granularity refers to how fine- / coarse-grained an answer is.\\

Always assign exactly one score: 1, 2, 3, or 4.\\

---

\{examples\}

---\\

Now assign a granularity score. Output only the score.\\

Answer: "\{answer\}"\\

Granularity:
\end{tcolorbox}

\section{LightGBM Model Training}
\label{app:model-params}
We train Granuscore using a LightGBM regression model on \textsc{GRANOLA-EQ}. Hyperparameters are selected with Optuna \citep{akiba_optuna_2019}, using 50 optimization trials on the development split. The final hyperparameter configuration is shown in \autoref{tab:lgb_hparams}.

\begin{table}[t]
\centering
\small
\begin{tabular}{ll}
\toprule
\textbf{Parameter} & \textbf{Value} \\
\midrule
Boosting type & GBDT \\
Objective & Regression \\
Evaluation metric & RMSE \\
Early stopping & 200 \\
Number of iterations & 10{,}000 \\
Learning rate & 0.0257596 \\
Number of leaves & 138 \\
Maximum depth & Unlimited \\
Minimum data in leaf & 57 \\
Feature fraction & 0.751449 \\
Bagging fraction & 0.638041 \\
Bagging frequency & 7 \\
Dropout rate (DART) & 0.1 \\
Maximum bins & 255 \\
Device & CPU \\
\bottomrule
\end{tabular}
\caption{LightGBM hyperparameters used for training Granuscore.}
\label{tab:lgb_hparams}
\end{table}

\section{Granola-EQ}
\label{app:granola-dataset}
\begin{table}[t]
\small
\centering
\label{tab:relation-questions}
\begin{tabular}{ll}
\toprule
Rel. & Question Template \\
\midrule
P112 & Who founded [X]? \\
P127 & Who owns [X]? \\
P131 & Where is [X] located? \\
P159 & Where is the headquarter of [X]? \\
P170 & Who created [X]? \\
P175 & Who performed [X]? \\
P176 & Which company is [X] produced by? \\
P19  & Where was [X] born? \\
P20  & Where did [X] die? \\
P26  & Who is [X] married to? \\
P264 & What music label represents [X]? \\
P276 & Where is [X] located? \\
P40  & Who is [X]'s child? \\
P50  & Who is the author of [X]? \\
P69  & Where was [X] educated? \\
P740 & Where was [X] founded? \\
\bottomrule
\end{tabular}
\caption{Question template for each relation type in the dataset.}
\label{tab:question-template}
\end{table}

\begin{table}[t]
\small
\centering
\begin{tabular}{lrrr}
\toprule
Rel. & Train & Dev & Test \\
\midrule
P20 & 650 & 36 & 49 \\
P19 & 635 & 45 & 64 \\
P69 & 600 & 48 & 30 \\
P276 & 585 & 28 & 55 \\
P159 & 496 & 76 & 67 \\
P26 & 482 & 58 & 142 \\
P131 & 452 & 96 & 61 \\
P176 & 444 & 73 & 124 \\
P50 & 443 & 64 & 104 \\
P170 & 439 & 105 & 56 \\
P264 & 398 & 68 & 95 \\
P127 & 361 & 83 & 147 \\
P40 & 337 & 83 & 127 \\
P112 & 235 & 38 & 57 \\
P175 & 79 & 315 & 35 \\
P740 & 66 & 4 & 7 \\
\midrule
Total & 6702 & 1220 & 1220 \\
\bottomrule
\end{tabular}
\caption{Distribution of relation types across the training, development, and test splits of \textsc{Granola-EQ}.}
\label{tab:relation-distribution}
\end{table}

\textsc{Granola-EQ} covers multiple relation types, represented through masked question templates. \autoref{tab:question-template} lists the relation categories included in the dataset together with their corresponding question templates. \autoref{tab:relation-distribution} reports the distribution of relation types across the training, development, and test splits. The dataset covers a diverse range of relations, with location- and person-based questions forming the largest categories. As shown in the table, the relative proportions differ across splits. This variation arises from enforcing a strict split-by-granularity-realization rule, ensuring that the same realization does not appear in multiple splits and preventing data leakage between training and evaluation.

\subsection{Distribution of Granularity Resolution}
\begin{table}[t]
\small
\centering
\begin{tabular}{l r}
\toprule
Gran. Resolution & Count \\
\midrule
1 & 206 \\
2 & 1966 \\
3 & 5650 \\
4 & 1320 \\
\midrule
Mean & 2.88 \\
Variance & 0.44 \\
\bottomrule
\end{tabular}
\caption{Distribution of granularity resolution in \textsc{Granola-EQ}. The granularity resolution indicates the number of distinct granularity levels available for an entity.}
\label{tab:granola-level-dist}
\end{table}

\autoref{tab:granola-level-dist} shows the distribution of granularity resolution across the full dataset. Granularity resolution denotes the number of levels available for a given entity. The mean and variance indicate that most entities exhibit a moderate number of granularity levels, with the majority containing three levels.

\subsection{Context-dependent granularity variation}
\begin{table}[t]
\small
\centering
\begin{tabular}{l r}
\toprule
Gran. Level & Count \\
\midrule
1 & 2 \\
2 & 13 \\
2.5 & 114 \\
3 & 160 \\
4 & 198 \\
\midrule
Mean & 3.25 \\
Variance & 0.44 \\
\bottomrule
\end{tabular}
\caption{Distribution of normalized granularity levels for the realization \emph{England} in \textsc{Granola-EQ}.}
\label{tab:granola-england}
\end{table}

Granularity is context-dependent: the same entity may occur at different levels of abstraction depending on the question. As described in \autoref{sec:dataset}, \textsc{GRANOLA-EQ} preserves these contextual variations. \autoref{tab:granola-england} illustrates this effect for the entity \emph{England}, showing that identical entities appear at multiple normalized granularity levels across dataset entries. Granuscore therefore reflects an aggregated, context-averaged view of granularity suitable for global analysis.

\subsection{Granuscore Across Annotated Granularity Levels}
\label{app:granuscore-labeled-levels}
\begin{table}[t]
\small
\centering
\begin{tabular}{l r r r r r}
\toprule
Level & 1 & 2 & 2.5 & 3 & 4 \\
\midrule
Percentile   & 28.54 & 47.33 & 57.27 & 64.12 & 77.29 \\
Raw   & 1.55 & 1.99 & 2.19 & 2.41 & 2.79 \\
\bottomrule
\end{tabular}
\caption{Average Granuscore per normalized granularity level on the \textsc{GRANOLA-EQ} test set using the HiT Random Anchor method. We report both the percentile score and the corresponding raw value.}
\label{tab:gran_level}
\end{table}

To examine whether Granuscore meaningfully distinguishes between granularity levels, \autoref{tab:gran_level} reports the average raw and percentile scores for each normalized level in the \textsc{Granola-EQ} test set.

Both the raw Granuscore values and their percentile equivalents increase consistently with the annotated granularity levels. The distances between adjacent levels are relatively uniform, indicating that Granuscore reflects the intended ordering of abstraction levels. This regular spacing suggests that Granuscore provides an interpretable scale of semantic granularity rather than producing small, arbitrary numerical differences. Pairwise comparisons between all levels are statistically significant (one-sided Wilcoxon signed-rank test; lower granularity $<$ higher granularity; $p \leq 6.4 \times 10^{-6}$).

\subsection{Additional Metrics}
\label{app:additional_metrics_granola}
\begin{table}[t]
\small
\setlength{\tabcolsep}{2pt}
\centering
\begin{tabular}{l@{\hspace{0pt}}ccc}
\toprule
Method & Kendall $\tau$ & Pearson $r$ & \makecell{Intra\\ Kendall $\tau$} \\
\midrule
Word Count & 4.15 & 1.07 & 7.91 \\
WordNet$^\dagger$ & 25.54 & 28.15 & 61.15 \\
GPT-4.1 mini & \textbf{69.84} & \textbf{74.04} & \textbf{88.24} \\
HiT Dist0 & 50.70 & 52.37 & 75.73 \\
\midrule
MiniLM NN & 28.65 & 36.94 & 39.00 \\
MiniLM Random & 28.79 & 37.87 & 41.14 \\
MiniLM RandomAnch & 28.73 & 37.76 & 42.31 \\
\midrule
HiT NN & 52.28 & 65.13 & 72.93 \\
HiT Random & 50.53 & 64.41 & 69.49 \\
HiT RadialAnch & 53.16 & 66.04 & 76.70 \\
HiT RandomAnch & 54.28 & 67.58 & 76.30 \\
\midrule
HiT Dist0 + NN & 54.04 & 67.55 & 74.05 \\
HiT Dist0 + Random & 52.51 & 66.67 & 73.65 \\
HiT Dist0 + RadialAnch & 54.66 & 68.91 & 77.66 \\
HiT Dist0 + RandomAnch & \textit{55.53} & \textit{69.69} & \textit{78.06} \\
\bottomrule
\end{tabular}
\caption{Kendall's $\tau$, Pearson $r$, and Intra-sample Kendall's $\tau$ on \textsc{GRANOLA-EQ}. Kendall's $\tau$ measures global ordering across all answers, while Intra-sample Kendall's $\tau$ measures ordering within individual questions. $^\dagger$: WordNet could only derive a hierarchy for 17.61\% of the answers in the test set.}
\label{tab:granola-eq-others}
\end{table}

\autoref{tab:granola-eq-others} reports additional evaluation metrics on \textsc{Granola-EQ}, including Kendall's $\tau$, Pearson's $r$, and Intra-sample Kendall's $\tau$. 
Kendall's $\tau$ and Pearson's $r$ are computed across realizations from all samples, measuring how well predicted granularity scores follow the global ordering of annotated levels across unrelated entities. 
In contrast, Intra-sample Kendall's $\tau$ is computed within each question and then averaged, reflecting how well a method preserves the ordering of realizations within individual hierarchies.

Under these metrics, the LLM baseline (\textsc{GPT-4.1 mini}) achieves the highest scores, followed by HiT Random Anchor. However, this result should be interpreted with caution. 
The LLM frequently assigns identical granularity levels to multiple answers, resulting in many tied comparisons. 
Since Kendall's $\tau$ excludes tied pairs from the computation, these ties effectively remove more difficult comparisons and leave only easier ordering decisions, artificially inflating the score.

For this reason, Kendall-based metrics can overestimate the performance of models that produce many tied predictions. 
To provide a more faithful evaluation of hierarchy recovery, we therefore report Pairwise Accuracy and Exact Ordering Accuracy in \autoref{res-granola-eq}, which evaluate the ordering of all answer pairs and penalize tied predictions.

\section{Ablation Study on the Number of References}
\label{app:k-ablation}
\begin{table}[t]
\small
\centering
\begin{tabular}{lc}
\toprule
Reference Size & PW Acc. \\
\midrule
33 & 83.37 \\
66 & 83.16 \\
99 & 83.13 \\
333 & 83.30 \\
666 & 82.44 \\
999 & \textbf{83.76} \\
1332 & 82.47 \\
1665 & \textit{83.45} \\
\bottomrule
\end{tabular}
\setlength{\belowcaptionskip}{-10pt}
\caption{Ablation study on the number of reference anchors on \textsc{Granola-EQ} using Random Anchors. Bold indicates the best result, and italics the second-best result across anchor sizes.}
\label{tab:k-ablation}
\end{table}

We analyze the effect of the number of reference anchors when using the Random Anchor method. 
\autoref{tab:k-ablation} reports pairwise accuracy on the \textsc{Granola-EQ} test split while varying the anchor set size.

Performance remains largely stable across different anchor sizes, indicating that the method is not highly sensitive to this parameter. 
The best performance is obtained with 999 anchors, which we therefore use as the default configuration in our experiments.

\section{Ablation Study on Aggregation Strategy}
\label{app:aggregation_strategy}
Because large-scale annotations of granularity for longer texts are difficult to obtain, we use our scientific papers testbed to determine an effective aggregation strategy. We compare several aggregation operators, including \textit{mean}, \textit{weighted mean}, \textit{sum}, \textit{min}, \textit{max}, and \textit{lower quantile mean} (lqm). The lower quantile mean with threshold $q$ averages only the lowest $q$ proportion of unit-level scores within a section (e.g., $\text{lqm}(0.3)$ averages the lowest 30\% of scores).

\autoref{tab:aggregation_ablation_papers} reports ordering accuracy for the different aggregation strategies. The best performance is achieved by a two-step aggregation procedure. First, we compute a sentence-level Granuscore by averaging the scores of the extracted referential units within each sentence. We then aggregate across sentences by taking the mean of the lowest 80\% of sentence-level Granuscores, which reduces the influence of unusually high values.

Some aggregation variants perform substantially worse than others for methodological reasons. In particular, max-based strategies (e.g., \texttt{sent-max-pool-max}, \texttt{doc-pool-max}) reduce an entire document to a single referential unit, effectively ignoring most of the content. Since many sentences contain at least some coarse or vague elements, these methods systematically bias scores toward coarse-grained representations and therefore provide poor discrimination.

For sum-based strategies (e.g., \texttt{doc-pool-sum} and \texttt{sent-sum-*}), the issue is different: scores accumulate additively across sentences, causing document-level granularity estimates to scale with text length. This behavior conflicts with our notion of granularity, which is determined by the hierarchical level of referential expressions rather than the amount of information conveyed by a text.

For completeness, we additionally evaluate all aggregation strategies in generalized additive models predicting sentence specificity from length and granularity. \autoref{tab:aggregation_ablation_gams} reports the corresponding improvements in explained deviance. The rankings induced by ordering accuracy and explained deviance are moderately correlated (Pearson $r = 0.62$), indicating that aggregation strategies that better recover hierarchical ordering also tend to better explain sentence specificity.

\section{Correlation to Sentence Specificity}
\label{app:specificity}

\begin{figure}[t]
    \centering
    \includegraphics[width=\linewidth]{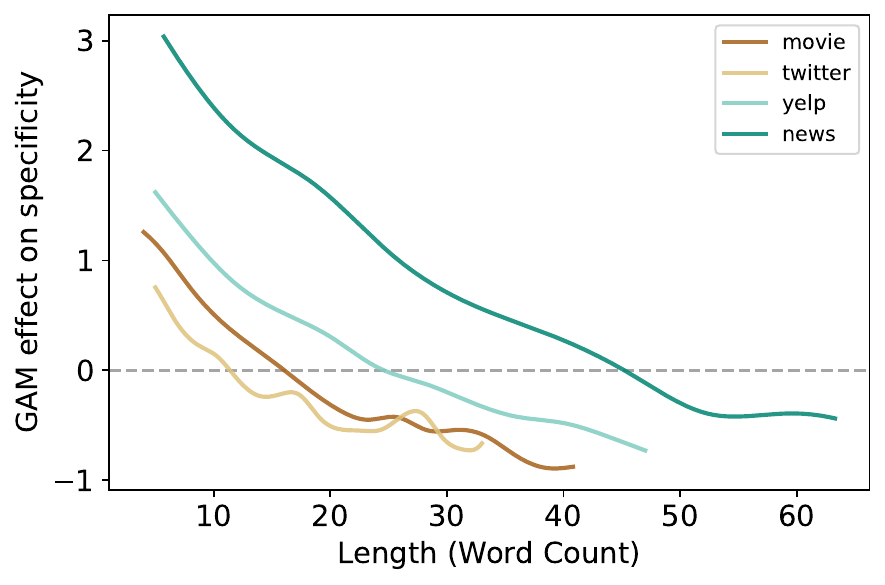}
\setlength{\belowcaptionskip}{-10pt}
\caption{Effect of Length on sentence specificity across domains.
Longer sentences correspond to more specific sentences.
The plotted range is restricted to the 1st--99th percentiles of Granuscore to avoid sparse-support regions.}
\label{fig:specificity-len}
\end{figure}

The sentence specificity datasets include 920 movie review sentences, 984 Twitter posts, and 845 Yelp reviews from \citet{ko_domain_2019}, as well as 573 news sentences from \citet{li_improving_2016}.

For completeness, \autoref{fig:specificity-gran} shows the estimated effect of sentence length (measured as word count) on sentence specificity across domains, restricted to the data-supported Granuscore range. As expected, we observe a negative relationship: as sentence length increases, specificity scores decrease, indicating more specific sentences.

The strength of this effect varies by domain. For Twitter, specificity decreases most rapidly with increasing length, followed by movie reviews, Yelp reviews, and news articles. This pattern reflects domain-specific length distributions: Twitter texts are typically much shorter than those in other domains, while news articles tend to be longer and more descriptive.

\section{Scientific Papers as Discourse Contexts}
\label{app:papers}
We compare paragraphs from the \textit{Introduction} and \textit{Related Work} sections. 
These sections are selected because their communicative roles are well defined: 
the Introduction typically presents the research problem and context, whereas the 
Related Work section situates the contribution within existing literature.

We sample the first 1{,}000 papers from the \textsc{S2ORC} corpus that contain standard \textit{Introduction}, \textit{Related Work}, and \textit{Conclusion} sections, ensuring a consistent and well-structured discourse layout. Before analysis, we remove bracketed text, URLs, figure captions, and common PDF/OCR artifacts.

To obtain comparable text segments across papers, we apply a simple paragraph 
selection procedure. For the Introduction, we select the first paragraph containing 
at least ten referential units. For the Related Work section, we skip the opening 
paragraph, as it often functions as a brief transition, and instead select the 
first subsequent paragraph meeting this criterion. If no such paragraph exists, 
we fall back to the opening paragraph if it also satisfies the requirement. 
This procedure yields 978 papers for comparison.

\section{Additional QA Analyses and Potential Confounding Factors}
\label{app:qa_datasets}
\begin{figure}[t]
    \includegraphics[width=\linewidth]{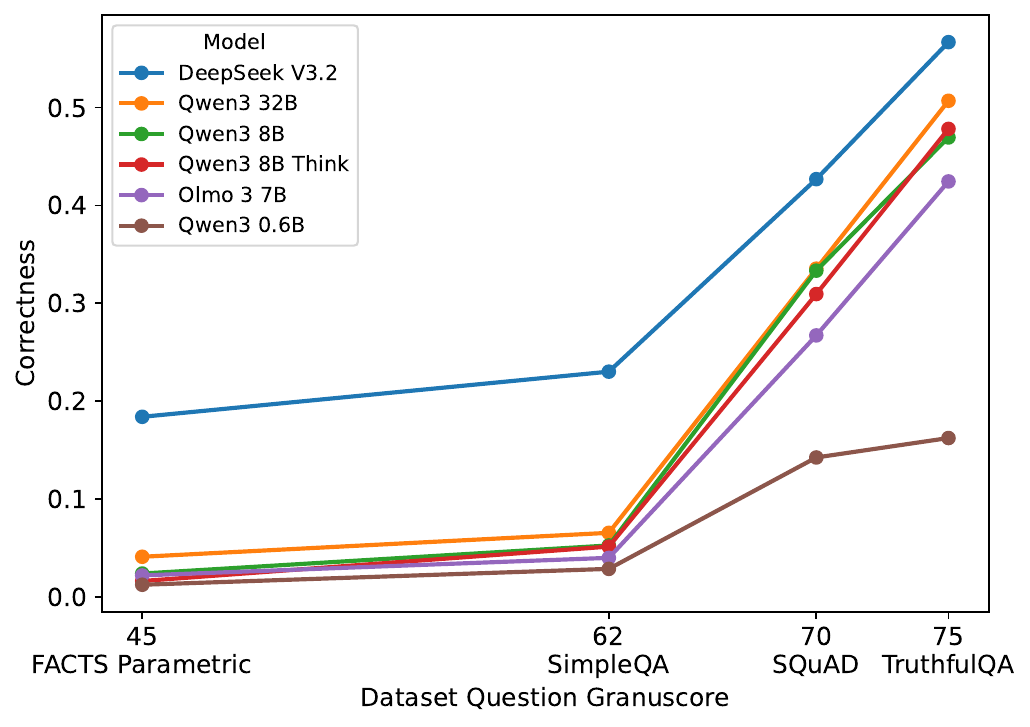}
    \caption{Relationship between dataset-level question Granuscore and model correctness across QA benchmarks. Higher Granuscore datasets are associated with higher correctness across models. All pairwise differences in Granuscore between datasets are statistically significant (Mann–Whitney $U$, $p \leq 3.2 \times 10^{-30}$).}
    \label{fig:qa_datasets_scatter-question}
\end{figure}

\paragraph{Question Granularity}
Beyond the correlation between the Granuscore of gold answers and model correctness reported in \autoref{sec:qa}, we also analyze the Granuscore of the corresponding questions. \autoref{fig:qa_datasets_scatter-question} shows a similar trend: datasets with lower question Granuscore are associated with lower correctness. This pattern is consistent across models. As for gold answers (\autoref{fig:qa_datasets_scatter}), the smallest model (\textsc{Qwen3 0.6B}) exhibits a weaker slope and partial saturation, whereas \textsc{DeepSeek V3.2} follow the same overall trend but at higher accuracy levels.

\paragraph{Potential Confounding Factors}
\begin{table*}[t]
\centering
\small
\begin{tabular}{
l
S[table-format=1.3]
S[table-format=1.3]
S[table-format=2.3]
S[table-format=1.3]
S[table-format=1.3]
S[table-format=2.3]
S[table-format=2.3]
}
\toprule
& & \multicolumn{2}{c}{Word Frequency} & \multicolumn{2}{c}{Tree Depth} & \multicolumn{2}{c}{Length} \\
\cmidrule(lr){3-4}
\cmidrule(lr){5-6}
\cmidrule(lr){7-8}
Dataset & {Accuracy} & {Answer} & {Question} & {Answer} & {Question} & {Answer} & {Question} \\
\midrule
FACTS Parametric & 0.050 & 0.018 & 0.153 & 2.265 & 3.518 & 3.108 & 6.198 \\
SimpleQA         & 0.078 & 0.022 & 0.510 & 1.859 & 6.315 & 2.240 & 16.310 \\
SQuAD            & 0.302 & 0.062 & 0.662 & 2.249 & 4.939 & 2.957 & 10.198 \\
TruthfulQA       & 0.435 & 1.051 & 0.808 & 4.244 & 4.775 & 9.118 & 10.620 \\
\bottomrule
\end{tabular}
\caption{Dataset-level statistics for alternative properties potentially related to QA difficulty. Correctness corresponds to mean model correctness across evaluated models.}
\label{tab:qa_confounds}
\end{table*}

We additionally analyze several alternative properties potentially related to QA difficulty: answer and question length, word frequency, and syntactic complexity. For word frequency, we compute the average token frequency using \texttt{wordfreq}~\citep{robyn_speer_2018_1443582}. For syntactic complexity, we measure average dependency tree depth using spaCy parses~\citep{honnibal_spacy_2020}. \autoref{tab:qa_confounds} reports the corresponding dataset-level statistics.

We observe a mild relationship between correctness and word frequency, with lower-performing datasets generally containing rarer terms. However, this effect is also partially related to granularity itself, since fine-grained concepts tend to be less frequent. In contrast, syntactic complexity and length-based measures do not exhibit comparably consistent relationships with correctness. Together, these findings suggest that the observed Granuscore trends are not explained solely by superficial textual properties.

\section{QA Generation and Evaluation}
\label{app:qa-generation}
For answer generation, we use a maximum length of 512 tokens for standard and 2{,}048 tokens for reasoning-based generation, with the temperature set to 0. We instruct the model to produce answers of at most five sentences. We retain only responses that terminate before the token limit, ensuring all evaluated outputs are complete and not truncated.

Models are instructed to produce answers of at most five sentences using the following prompt:

\begin{tcolorbox}[
  colback=bluebg,
  colframe=blue!60!black,
  coltitle=white,
  fonttitle=\bfseries,
  fontupper=\small\ttfamily,
  boxrule=0.5mm,
  rounded corners,
  title={User Prompt: Answer Generation}
]
Answer the following query in at most five complete sentences: <query>
\end{tcolorbox}

Model responses are evaluated using GPT-4.1 nano as an LLM-based judge, following the prompt template introduced in \textsc{SimpleQA} \citep{wei_measuring_2024}.

\begin{table*}[t]
    \centering
    \small
\begin{tabular}{lclclc}
\toprule
Aggregation & Ord. Acc. & Aggregation & Ord. Acc. & Aggregation & Ord. Acc. \\
\midrule
sent-weighted-mean-pool-sum & 58.49 & sent-lqm-0.9-pool-sum & 60.84 & sent-min-pool-sum & 60.02 \\
sent-weighted-mean-pool-mean & 66.46 & sent-lqm-0.8-pool-sum & 62.27 & sent-min-pool-mean & 66.16 \\
sent-weighted-mean-pool-lqm-0.1 & 62.68 & sent-lqm-0.7-pool-sum & 62.68 & sent-min-pool-lqm-0.1 & 61.76 \\
sent-weighted-mean-pool-lqm-0.3 & 66.87 & sent-lqm-0.9-pool-mean & \textit{68.10} & sent-min-pool-lqm-0.3 & 64.01 \\
sent-weighted-mean-pool-lqm-0.5 & 67.28 & sent-lqm-0.8-pool-mean & \textbf{68.71} & sent-min-pool-lqm-0.5 & 65.03 \\
sent-weighted-mean-pool-min & 61.55 & sent-lqm-0.7-pool-mean & 67.59 & sent-min-pool-min & 59.82 \\
sent-weighted-mean-pool-max & 58.79 & sent-lqm-0.9-pool-lqm-0.1 & 63.80 & sent-min-pool-max & 60.53 \\
sent-sum-pool-sum & 47.55 & sent-lqm-0.9-pool-lqm-0.3 & 66.16 & sent-max-pool-sum & 52.66 \\
sent-sum-pool-mean & 45.19 & sent-lqm-0.9-pool-lqm-0.5 & 66.87 & sent-max-pool-mean & 54.19 \\
sent-sum-pool-lqm-0.1 & 48.67 & sent-lqm-0.8-pool-lqm-0.1 & 63.80 & sent-max-pool-lqm-0.1 & 53.58 \\
sent-sum-pool-lqm-0.3 & 47.03 & sent-lqm-0.8-pool-lqm-0.3 & 67.59 & sent-max-pool-lqm-0.3 & 54.91 \\
sent-sum-pool-lqm-0.5 & 45.71 & sent-lqm-0.8-pool-lqm-0.5 & 67.89 & sent-max-pool-lqm-0.5 & 56.85 \\
sent-sum-pool-min & 48.67 & sent-lqm-0.7-pool-lqm-0.1 & 63.29 & sent-max-pool-min & 52.56 \\
sent-sum-pool-max & 44.38 & sent-lqm-0.7-pool-lqm-0.3 & 67.38 & sent-max-pool-max & 48.47 \\
sent-mean-pool-sum & 60.53 & sent-lqm-0.7-pool-lqm-0.5 & 67.59 & doc-pool-sum & 47.55 \\
sent-mean-pool-mean & 67.48 & sent-lqm-0.9-pool-min & 62.27 & doc-pool-mean & 66.36 \\
sent-mean-pool-lqm-0.1 & 62.88 & sent-lqm-0.8-pool-min & 62.07 & doc-pool-lqm-0.1 & 64.83 \\
sent-mean-pool-lqm-0.3 & 66.05 & sent-lqm-0.7-pool-min & 62.07 & doc-pool-lqm-0.3 & 65.64 \\
sent-mean-pool-lqm-0.5 & 66.16 & sent-lqm-0.9-pool-max & 59.41 & doc-pool-lqm-0.5 & 66.05 \\
sent-mean-pool-min & 62.07 & sent-lqm-0.8-pool-max & 60.33 & doc-pool-min & 59.92 \\
sent-mean-pool-max & 59.10 & sent-lqm-0.7-pool-max & 59.82 & doc-pool-max & 47.75 \\
\bottomrule
\end{tabular}
    \caption{Accuracy of section ordering (Introduction $>$ Related Work) under different aggregation strategies. 
    Aggregation names follow the pattern \texttt{scope-aggregation-pool}. 
    \texttt{scope} indicates whether aggregation is performed at the document level (\texttt{doc}) or sentence level (\texttt{sent}). 
    For sentence-level strategies, the first operator aggregates scores across sentences (e.g., \texttt{sent-mean}). 
    The \texttt{pool} operator specifies how Granuscores of referential units within a sentence are combined (e.g., \texttt{sent-mean-pool-sum} first sums Granuscores within each sentence and then averages across sentences). Bold and italics denote the best and second-best results, respectively.}
    \label{tab:aggregation_ablation_papers}
\end{table*}

\begin{table*}[t]
    \centering
    \small
    \setlength{\tabcolsep}{3pt}
\begin{tabular}{lclclc}
\toprule
Method & $\Delta$ Expl. Dev. & Method & $\Delta$ Expl. Dev. & Method & $\Delta$ Expl. Dev. \\
\midrule
sent-sum-pool-sum & 3.67 & sent-lqm-0.9-pool-lqm-0.3 & 9.20 & sent-max-pool-sum & 2.74 \\
sent-sum-pool-mean & 6.75 & sent-lqm-0.9-pool-lqm-0.5 & 9.20 & sent-max-pool-mean & 7.09 \\
sent-sum-pool-lqm-0.1 & 6.64 & sent-lqm-0.8-pool-lqm-0.1 & 9.20 & sent-max-pool-lqm-0.1 & 7.81 \\
sent-sum-pool-lqm-0.3 & 7.27 & sent-lqm-0.8-pool-lqm-0.3 & 9.24 & sent-max-pool-lqm-0.3 & 7.68 \\
sent-sum-pool-lqm-0.5 & 7.12 & sent-lqm-0.8-pool-lqm-0.5 & 9.24 & sent-max-pool-lqm-0.5 & 7.71 \\
sent-sum-pool-min & 6.10 & sent-lqm-0.7-pool-lqm-0.1 & 9.23 & sent-max-pool-min & 7.28 \\
sent-sum-pool-max & 1.51 & sent-lqm-0.7-pool-lqm-0.3 & 9.35 & sent-max-pool-max & 2.00 \\
sent-mean-pool-sum & 2.55 & sent-lqm-0.7-pool-lqm-0.5 & 9.29 & doc-pool-sum & 3.49 \\
sent-mean-pool-mean & 8.34 & sent-lqm-0.9-pool-min & 8.76 & doc-pool-mean & 8.56 \\
sent-mean-pool-lqm-0.1 & 9.19 & sent-lqm-0.8-pool-min & 8.77 & doc-pool-lqm-0.1 & \textit{9.84} \\
sent-mean-pool-lqm-0.3 & 9.20 & sent-lqm-0.7-pool-min & 8.78 & doc-pool-lqm-0.3 & 9.52 \\
sent-mean-pool-lqm-0.5 & 9.20 & sent-lqm-0.9-pool-max & 1.87 & doc-pool-lqm-0.5 & 9.48 \\
sent-mean-pool-min & 8.76 & sent-lqm-0.8-pool-max & 1.86 & doc-pool-min & 9.51 \\
sent-mean-pool-max & 1.87 & sent-lqm-0.7-pool-max & 1.85 & doc-pool-max & 1.88 \\
sent-lqm-0.9-pool-sum & 2.55 & sent-min-pool-sum & 1.95 & sent-weighted-mean-pool-sum & 2.65 \\
sent-lqm-0.8-pool-sum & 2.56 & sent-min-pool-mean & 8.07 & sent-weighted-mean-pool-mean & 8.62 \\
sent-lqm-0.7-pool-sum & 2.50 & sent-min-pool-lqm-0.1 & \textbf{9.98} & sent-weighted-mean-pool-lqm-0.1 & 9.14 \\
sent-lqm-0.9-pool-mean & 8.34 & sent-min-pool-lqm-0.3 & 9.42 & sent-weighted-mean-pool-lqm-0.3 & 9.41 \\
sent-lqm-0.8-pool-mean & 8.33 & sent-min-pool-lqm-0.5 & 9.18 & sent-weighted-mean-pool-lqm-0.5 & 9.54 \\
sent-lqm-0.7-pool-mean & 8.33 & sent-min-pool-min & 9.53 & sent-weighted-mean-pool-min & 8.69 \\
sent-lqm-0.9-pool-lqm-0.1 & 9.19 & sent-min-pool-max & 1.84 & sent-weighted-mean-pool-max & 1.90 \\
\bottomrule
\end{tabular}
\caption{Ablation over aggregation strategies measured by the improvement in explained deviance ($\Delta$ Expl. Dev., $\times 100$) relative to a length-only baseline for sentence specificity. 
Aggregation names follow the pattern \texttt{scope-aggregation-pool}. 
\texttt{scope} indicates whether aggregation is performed at the document level (\texttt{doc}) or sentence level (\texttt{sent}). 
For sentence-level strategies, the first operator aggregates scores across sentences, while \texttt{pool} specifies how Granuscores of referential units within each sentence are combined. 
For example, \texttt{sent-mean-pool-sum} first sums Granuscores within each sentence and then averages across sentences. 
Bold and italics denote the best and second-best results, respectively.}
    \label{tab:aggregation_ablation_gams}
\end{table*}

\end{document}